\documentclass[10pt,twocolumn,letterpaper]{article}

\usepackage{cvpr}
\usepackage{times}
\usepackage{epsfig}
\usepackage{graphicx}
\usepackage{amsmath}
\usepackage{amssymb}
\usepackage{multirow}
\usepackage{booktabs} 
\usepackage{algorithm}
\usepackage{algorithmic}

\usepackage{adjustbox}
\usepackage{booktabs}
\usepackage{multirow}
\usepackage{subcaption}
\usepackage{tabularx}
\usepackage{float}

\usepackage[pagebackref=true,breaklinks=true,letterpaper=true,colorlinks,bookmarks=false]{hyperref}

\cvprfinalcopy 

\def\cvprPaperID{263} 

\ifcvprfinal\fi
\begin{document}

\title{Recurrent 3D Pose Sequence Machines}

\author{Mude Lin, \quad Liang Lin\thanks{Corresponding author is Liang Lin. This work was supported by State Key Development Program under Grant 2016YFB1001004, NSFC-Shenzhen Robotics Projects(U1613211), and the Fundamental Research Funds for the Central Universities, and Guangdong Science and Technology Program under Grant 201510010126.}, \quad Xiaodan Liang, \quad Keze Wang, \quad Hui Cheng\\
School of Data and Computer Science, Sun Yat-sen University\\
{\tt\small linmude@foxmail.com}, {\tt\small linliang@ieee.org}, {\tt\small xdliang328@gmail.com},\\{\tt\small wangkeze@mail2.sysu.edu.cn}, {\tt\small chengh9@mail.sysu.edu.cn}
}

\maketitle

\begin{abstract}3D human articulated pose recovery from monocular image sequences is very challenging due to the diverse appearances, viewpoints, occlusions, and also the human 3D pose is inherently ambiguous from the monocular imagery. It is thus critical to exploit rich spatial and temporal long-range dependencies among body joints for accurate 3D pose sequence prediction. Existing approaches usually manually design some elaborate prior terms and human body kinematic constraints for capturing structures, which are often insufficient to exploit all intrinsic structures and not scalable for all scenarios. In contrast, this paper presents a Recurrent 3D Pose Sequence Machine(RPSM) to automatically learn the image-dependent structural constraint and sequence-dependent temporal context by using a multi-stage sequential refinement. At each stage, our RPSM is composed of three modules to predict the 3D pose sequences based on the previously learned 2D pose representations and 3D poses: (i) a 2D pose module extracting the image-dependent pose representations, (ii) a 3D pose recurrent module regressing 3D poses and (iii) a feature adaption module serving as a bridge between module (i) and (ii) to enable the representation transformation from 2D to 3D domain. These three modules are then assembled into a sequential prediction framework to refine the predicted poses with multiple recurrent stages. Extensive evaluations on the Human3.6M dataset and HumanEva-I dataset show that our RPSM outperforms all state-of-the-art approaches for 3D pose estimation.
\end{abstract}


\section{Introduction}

\begin{figure}[h]
\centering
\includegraphics[width=0.99 \columnwidth]{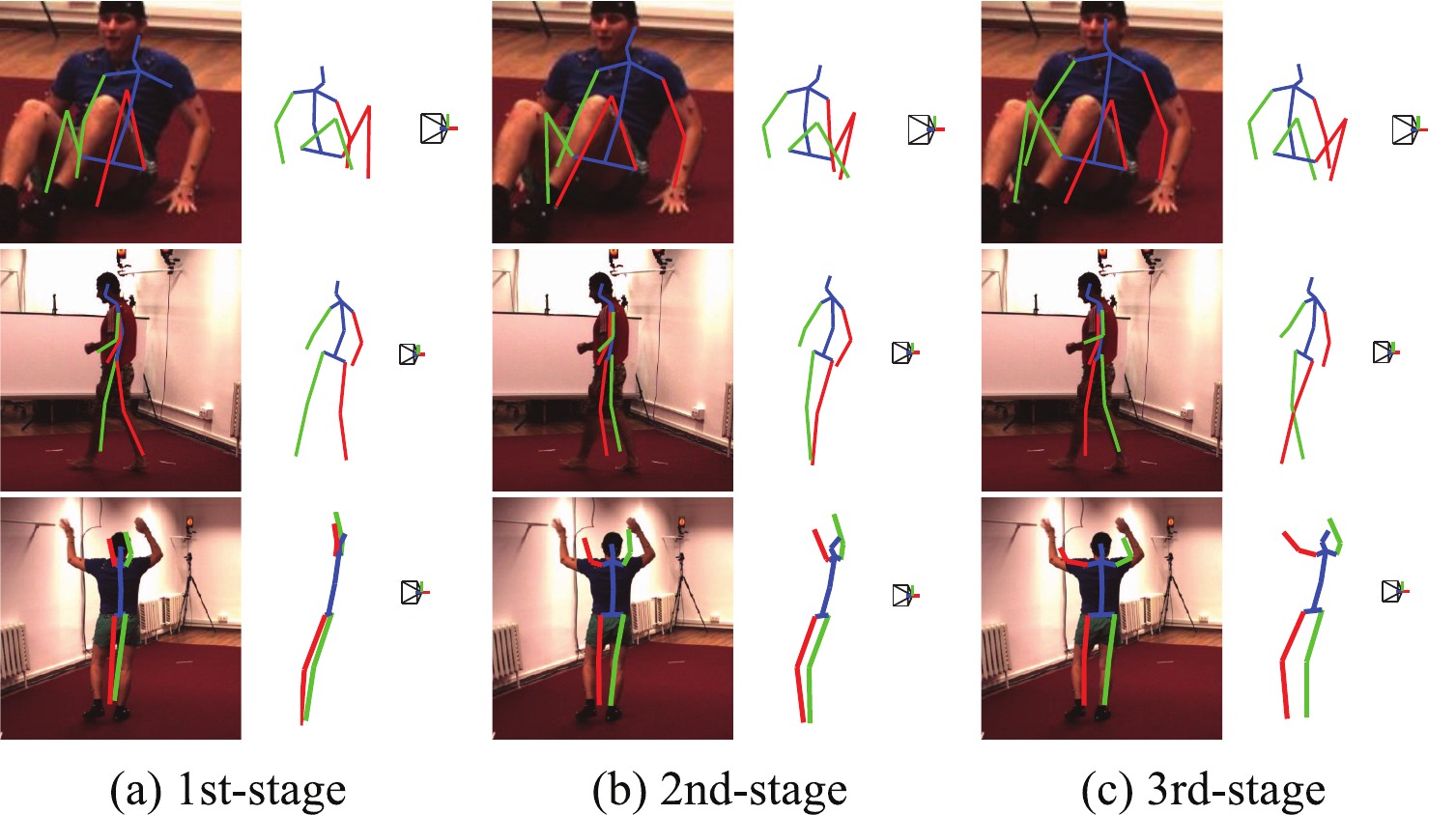}
\caption{Some visual results of our approach (RPSM) on Human3.6M dataset. The estimated 3D skeletons are reprojected into the images and shown by themselves from the side view (next to the images). The figures from left to right correspond to the estimated 3D poses generated by the 1st-stage, 2nd-stage and 3rd-stage of RPSM, respectively. We can observe that the predicted human 3D joints are progressively corrected along with the multi-stage sequential learning. Best viewed in color.}
\label{fig:3d_pose_example}
\end{figure}

Though quite challenging, recovering the 3D full-body human pose from a monocular RGB image sequence has recently attracted a lot of research interests due to its huge potentials on high-level applications, which includes human-computer interaction~\cite{errity2016human}, surveillance~\cite{held2012intelligent}, video browsing/indexing~\cite{chen2013video} and virtual reality~\cite{zz_rheingold1991virtual}.

Besides the challenges shared with 2D image pose estimation (e.g., large variation in human appearance, arbitrary camera viewpoints and obstructed visibilities due to external entities and self-occlusions), 3D articulated pose recovery from monocular imagery is much more difficult since 3D pose is inherently ambiguous from a geometric perspective~\cite{zhou2016deep}, as shown in Fig.~\ref{fig:3d_pose_example}. To resolve all these issues, a preferable way is to investigate how to simultaneously enforce 2D spatial relationship, 3D geometry constraint and temporal consistency within one single model.

Recently, notable successes have been achieved for 2D pose estimation based on 2D part models coupled with 2D deformation priors, \eg, \cite{xiaohan2015joint,yang2011articulated}, and the deep learning techniques, \eg, \cite{deeppose, deepfacialpoint, wei2016convolutional,yang2016end}. However, these methods have not explored the 3D pose geometry that is crucial for 3D pose estimation. There has been some limited attempts on combining the image-based 2D part detectors, 3D geometric pose priors and temporal models for generating 3D poses \cite{andriluka2010monocular,zhou2014spatio,zhou2015sparseness,Tekin_2016_CVPR}. They mainly follow two kinds of pipelines: the first \cite{zhou2015sparseness,li2015maximum} resorts to the model-based 3D pose reconstruction by using external 3D pose gallery, while the second pipeline \cite{bogo2016keep,zhou2016deep} focuses on elaborately designing human body kinematic constraints with the model training. These separate techniques and prior knowledge make their models very sophisticated. Hence, validating the effectiveness of their each component is also not straightforward. In contrast to all these mentioned methods, we introduce a completely data-driven approach that learns to integrate the 2D spatial relationship, 3D geometry and temporal smoothness for the network training in a fully differential way.

We propose a novel Recurrent 3D Pose Sequence Machine (RPSM) for estimating 3D human poses from a sequence of images. Inspired by the pose machine \cite{ramakrishna2014pose} and convolutional pose machine \cite{wei2016convolutional} architectures for 2D pose estimation, our RPSM proposes a multi-stage training to capture long-range dependencies among multiple body-parts for 3D pose prediction, and further enforce the temporal consistency between the predictions of sequential frames. Specifically, the proposed RPSM recursively refines the predicted 3D pose sequences by sensing what already achieved in the previous stages, i.e., 2D pose representations and previously predicted 3D poses. At each stage, our RPSM is composed by a 2D pose module, a feature adaption module, and a 3D pose recurrent module. These three modules are constructed by the integration of the advanced convolutional and recurrent neural networks to fully exploit spatial and temporal constraints, which makes our RPSM with multi-stages a differentiable architecture that can be trained in an end-to-end way.


As illustrated in Fig.~\ref{fig:3d_pose_example}, our RPSM enables to gradually refine the 3D pose prediction for each frame with multiple sequential stages, contributing to seamlessly learning the image-dependent constraint between multiple body parts and sequence-dependent context from the previous frames. Specifically, at each stage, the 2D pose module takes each frame and 2D feature maps produced in previous stages as inputs and progressively updates the 2D pose representations. Then a feature adaption module is injected to transform learned pose representations from 2D to 3D domain. The 3D pose recurrent module, constructed by a Long-Short Term Memory (LSTM) layer, can thus regress the 3D pose estimation by combining the three lines of information, \ie the transformed 2D pose representations, 3D joint prediction from the previous stage and the memorized states from past frames. Intuitively, the 2D pose representations are conditioned on the monocular image which captures the spatial appearance and context information. 
The 3D joint prediction implicitly encodes the 3D geometry structural information by aggregating multi-stage computation. Then temporal contextual dependency is captured by the hidden states of LSTM units, which effectively improves robustness of the 3D pose estimations over time.


The main \textbf{contribution} of this work is three-fold. i) We propose a novel RPSM model that learns to recurrently integrate rich spatial and temporal long-range dependencies using a multi-stage sequential refinement, instead of relying on specifically manually defined body smoothness or kinematic constraints. ii) Casting the recurrent network models to sequentially incorporate 3D pose geometry structural information is innovative in literature, which may also inspire other 3D vision tasks. iii) Extensive evaluations on the public challenging Human3.6M dataset \cite{huamn3.6m} and HumanEva-I dataset \cite{sigal2010humaneva} show that our approach outperforms existing methods of 3D human pose estimation by large margins.

\section{Related work}
Considerable research has addressed the challenge of 3D human pose estimation. Early research on 3D monocular pose estimation from videos involves frame-to-frame pose tracking and dynamic models that rely on Markov dependencies among previous frames, \eg. \cite{wang2014robust,sigal2012loose}. The main drawbacks of these approaches are the requirement of the initialization pose and the inability to recover from tracking failure. To overcome these drawbacks, more recently approaches \cite{andriluka2010monocular,BurgosArtizzuBMVC13PoseNms} focus on detecting candidate poses in each individual frames and a post-processing step attempts to establish temporal consistent poses. Yasin \etal \cite{yasin2016dual} proposed a dual-source approach for 3D pose estimation from a single image. They combined the 3D pose data from motion capture system with image source annotated with 2D pose. They transformed the estimation to a 3D pose retrieval problem. One major limitation of this approach is the time efficiency. It takes more than 20 seconds to process an image. Sanzari \etal \cite{DBLP:conf/eccv/SanzariNP16} proposed a hierarchical Bayesian non-parametric model, which relies on a representation of the idiosyncratic motion of human skeleton joints groups and the consistency of the connected group poses is taken into account when reconstructing the full-body pose. Their approach achieved state-of-the-art performance on the Human3.6M~\cite{huamn3.6m} dataset.

\begin{figure*}[ht]
\centering
	\includegraphics[width=0.9\linewidth]{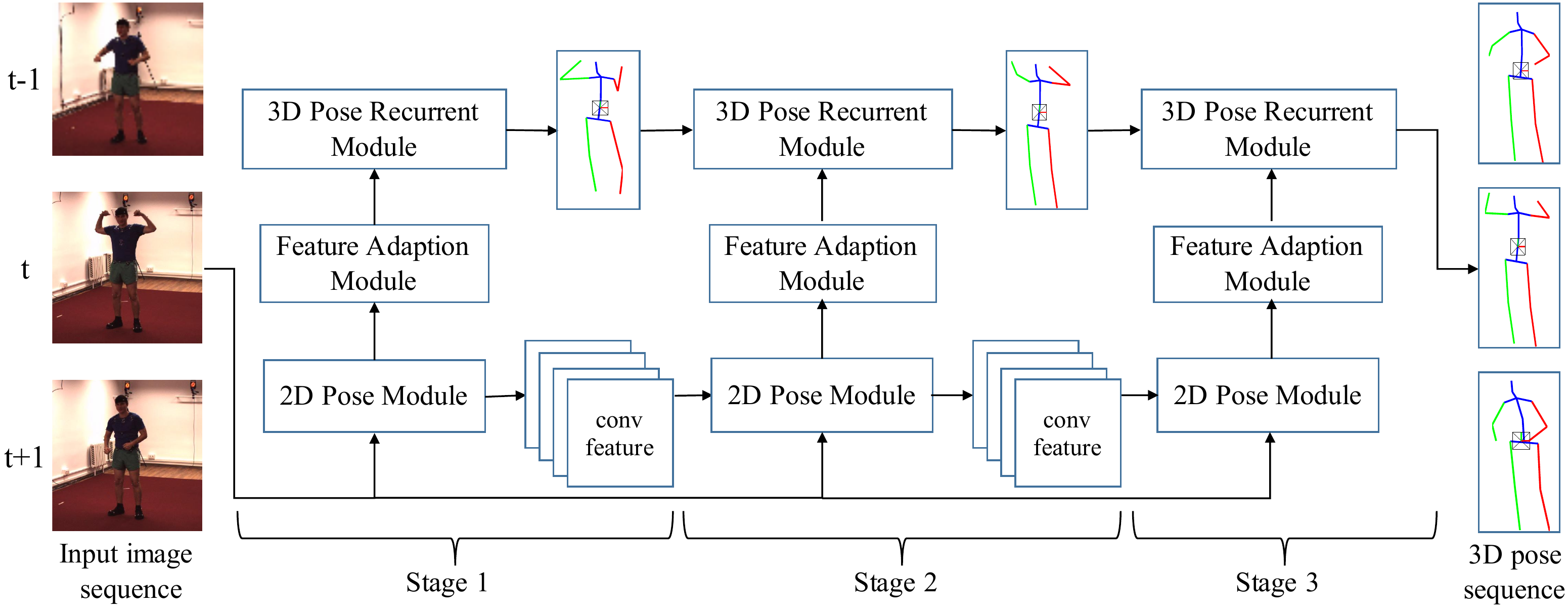}
	\vspace{-10pt}
	\caption{An overview of the proposed Recurrent 3D Pose Sequence Machine architecture. Our framework predicts the 3D human poses for all of the monocular image frames, and then sequentially refines them with multi-stage recurrent learning. At each stage, every frame of the input sequence is sequentially passed into three neural network modules: i) a 2D pose module extracting the image-dependent pose representations; 2) a feature adaption module for transforming the pose representations from 2D to 3D domain; 3) a 3D pose recurrent module predicting the human joints in 3D coordinates. Note that, the parameters of 3D pose recurrent module for all frames are shared to preserve the temporal motion coherence. Given the initial predicted 3D joints and 2D features from the first stage, we perform the multi-stage refinement to recurrently improve the pose accuracy. From the second stage, the previously predicted 17 joints (51 dimensions) and the 2D pose-aware features are also posed as the input of 2D pose module and 3D pose recurrent module, respectively. The final 3D pose sequence results are obtained after recurrently performing the multi-stage refinement. }
	\label{fig:framework}
	\vspace{-12pt}
\end{figure*}

Recently, deep learning has proven its ability in many computer vision tasks, such as the 3D human pose estimation. Li and Chan \cite{li20143d} firstly used the CNNs to regress the 3D human pose from monocular images and proposed two training strategies to optimize the network. Li \etal \cite{li2015maximum} proposed to integrate the structure-learning into deep learning framework, which consists of a convolutional neural network to extract image feature, and two following subnetworks to transform the image features and pose into a joint embedding. Tekin \etal \cite{Tekin_2016_CVPR} proposed to exploit motion information from consecutive frames and applied a deep learning network to regress the 3D pose. Zhou \etal \cite{zhou2015sparseness} proposed a 3D pose estimation framework from videos that consists of a novel synthesis between a deep-learning-based 2D part detector, a sparsity-driven 3D reconstruction approach and a 3D temporal smoothness prior. Zhou \etal \cite{zhou2016deep} proposed to directly embed a kinematic object model into the deep learning. Du \etal \cite{DBLP:conf/eccv/DuWLHGWKG16} introduced an additional built-in knowledge for reconstructing the 2D pose and formulated a new objective function to estimate 3D pose from detected 2D pose.

\section{Recurrent 3D Pose Sequence Machines}
As illustrated in Fig.~\ref{fig:framework}, we propose a novel Recurrent 3D Pose Sequence Machine (RPSM) to resolve 3D pose sequence generation for monocular frames, which recurrently refines the predicted 3D poses at multiple stages. At each stage, RPSM consists of three consecutive modules: 1) 2D pose module to extracts 2D pose-aware features; 2) feature adaption module to transform the representation from 2D to 3D domain; 3) 3D pose recurrent module to estimate 3D poses for each frame incorporating temporal dependency in the image sequence. These three modules are combined into a unified framework in each stage. The monocular image sequences are passed into multiple stages to gradually refine the predicted 3D poses. We train the network parameters recurrently at multiple stages in a fully end-to-end way.

\subsection{Multi-stage Optimization}
The 3D human pose is often represented as a set of $P$ joints with 3D location relative to a root joint (\eg, pelvis joint). Some exemplar poses are shown in Fig.~\ref{fig:3d_pose_example}. Our goal is to learn a mapping function that predicts the 3D pose sequence $\{S_1,...,S_T\}$ for the image sequence $\{I_1,...,I_T\}$, where $I_t$ is the $t$-th frame containing a subject and $S_t \in \mathbb{R}^{3\times{P}}$ is its corresponding 3D joint locations. 

\begin{figure*}[ht]
	\centering
	\includegraphics[width=0.91 \textwidth]{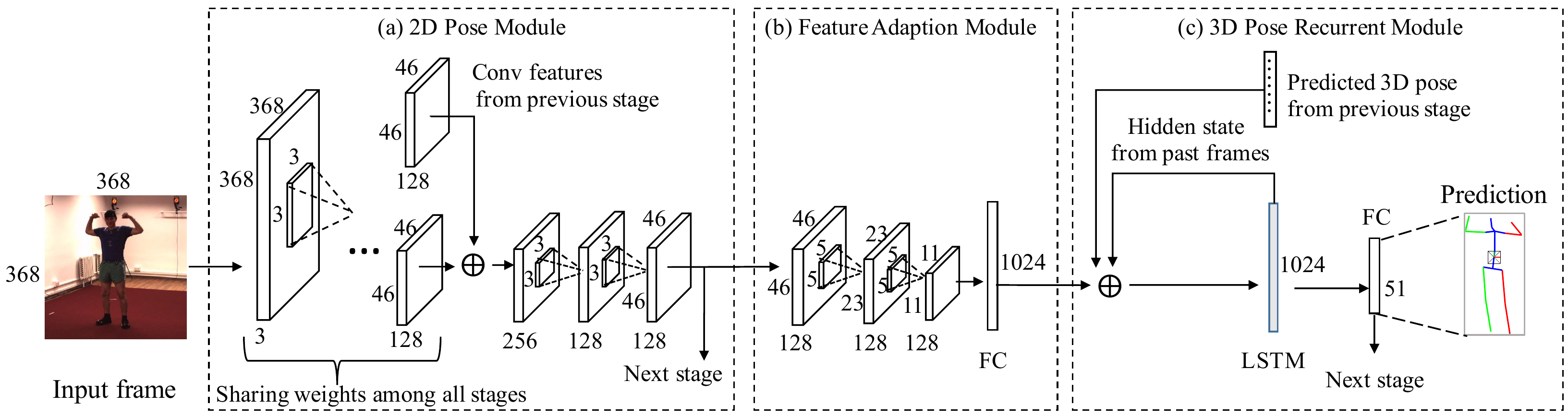}
	\vspace{-10pt}
	\caption{Detailed network architecture of our proposed RPSM at the $k$-th stage. An input frame with the $368\times368$ size is subsequently fed into 2D pose module, feature adaption module and 3D pose recurrent module to predict the locations of 17 joint points (51 dimensions output). The 2D pose module consists of $15$ shared convolution layers across all stages and $2$ specialized convolution layers for each stage. The specialized convolution layers take the shared features and the 2D pose-aware features at previous stage as the input, and output specialized features to the feature adaption module as well as the next stage. The feature adaption module consists of two convolution layers and one fully-connected layer with 1024 units. Finally, the adapted features, the hidden states of the LSTM layer and previously predicted 3D poses are concatenated together as the input of 3D pose recurrent module to produce the 3D pose of each frame. The symbol $\oplus$ means the concatenation operation.}
	\label{fig:stage_detail}
	\vspace{-10pt}
\end{figure*}

\begin{table*}
\label{fig:2D_network_shared}
\begin{adjustbox}{max width=1.0\textwidth}
\begin{tabular}{|c|c|c|c|c|c|c|c|c|c|}
\hline
 & 1 & 2 & 3 & 4 & 5 & 6 & 7 & 8 & 9 \\ \hline
Layer Name & conv1\_1 & conv1\_2 & max\_1 & conv2\_1 & conv2\_2 & max\_2 & conv3\_1 & conv3\_2 & conv3\_3 \\ \hline
Channel (kernel-stride) & 64(3-1) & 64(3-1) & 64(2-2) & 128(3-1) & 128(3-1) & 128(2-2) & 256(3-1) & 256(3-1) & 256(3-1) \\ \hline
 & 10 & 11 & 12 & 13 & 14 & 15 & 16 & 17 & 18 \\ \hline
Layer Name & conv3\_4 & max\_3 & conv4\_1 & conv4\_2 & conv4\_3 & conv4\_4 & conv4\_5 & conv4\_6 & conv4\_7 \\ \hline
Channel (kernel-stride) & 256(3-1) & 256(2-2) & 512(3-1) & 512(3-1) & 256(3-1) & 256(3-1) & 256(3-1) & 256(3-1) & 128(3-1) \\ \hline
\end{tabular}
\end{adjustbox}
\vspace{-10pt}
\caption{Details of the shared convolutional layers in 2D pose module. }
\label{table:shared_network_details}
\vspace{-10pt}
\end{table*}


Aiming at obtaining the 3D pose $S_t^k$ of the $t$-th frame at $k$-th stage, 2D pose module $\Psi_p$ is first employed to extract the 2D pose-aware features $f_{2D}^{t,k}$ for each image by taking the image $I_t$ and the previously 2D pose-aware features $f_{2D}^{t,k-1}$ as the input. Then the extracted 2D pose-aware features $f_{2D}^{t,k}$ are fed into the feature adaption module $\Psi_a$ to generate adapted features $f_{3D}^{t,k}$. Finally, the 3D pose $S_t^{k}$ is predicted according to the input of 3D pose recurrent module $\Psi_r$, which is composed of $f_{3D}^{t,k}$, the previously predicted 3D pose $S_t^{k-1}$ and the hidden states $H_{t-1}^k$ learned from the past frames. Formally, the $f_{2D}^{t,k} $, $S_t^k$, $f_{3D}^{t,k}$ of the $t$-th stage at $k$-th stage are formulated as, 
\begin{equation}
	\begin{aligned}
	f_{2D}^{t,k} &= \Psi_p(I_t,f_{2D}^{t,k-1}; W_p), \\
	f_{3D}^{t,k} &= \Psi_a(f_{2D}^{t,k}; W_a), \\
	S_t^k &= \Psi_r(f_{3D}^{t,k}, H_{t-1}^k,S_t^{k-1}; W_r),
	\label{eq:opt}
	\end{aligned}	
\end{equation}
where $W_p$, $W_a$, $W_r$ are network parameters of $\Psi_p, \Psi_a, \Psi_r$, respectively. At the first stage, the $f_{2D}^{t,0}$, $S_t^0$ are set as zero of the same size with those of other stages, and $H_0^k$ is set to be a vector of zeros. The 3D pose sequence $\{S_1^K,S_2^K,\dots,S_T^K\}$ estimated by the last $K$-th stage stage is the final prediction. 
The sequential refinement procedure of our RPSM enables the gradually updating of the network status to better learn the mapping between the image sequence and 3D pose sequence.

\subsection{2D Pose Module}
\label{sec:2d_pose_cnn}
The goal of the 2D pose module is to encode each frame in the monocular sequence with a compact representation of the pose information, \eg the body shape of the human. As a matter of fact, the lower convolution layers often extract the common low-level information, which is a very basic representation of the human image. Hence, we divide our proposed 2D pose module into two parts: the shared convolution layers across all stages and specialized pose-aware convolution layers in each stage. The architecture of 2D pose module is illustrated in Fig.~\ref{fig:stage_detail}(a). 

The shared convolution layers, i.e., those before the concatenation operation shown in Fig.~\ref{fig:stage_detail}(a), consist of 15 convolutional layers and four max-pooling layer. The kernel size of all shared convolutional layers are set to $3\times3$, and the four max-pooling layers are set to have $2\times2$ kernel with a stride of 2. The numbers of channels for the shared convolution layers from left to right in Fig.~\ref{fig:stage_detail}(a) are 64, 64, 128, 128, 256, 256, 256, 256, 512, 512, 256, 256, 256, 256 and 128, respectively (please see Table~\ref{table:shared_network_details} for more details). Moreover, we append the Rectified Linear Unit(ReLU) layers on all convolution layers. 

Afterwards, the shared convolution features and the extracted 2D pose-aware features at the previous stage are concatenated and then fed into the last two convolution layers to generate the updated 2D pose-aware features in 2D pose module. By combining the previously learned 2D pose-aware features at previous stage, the discriminative capability of the extracted 2D pose-aware features can be gradually enhanced, leading to a better 3D pose prediction. The higher convolution layers (i.e., the last 2 convolution layers in Fig.~\ref{fig:stage_detail}(a)) of 2D pose module often capture more structure-sensitive information, which should be specialized for the refinement at each stage. Thus, we train the network parameters of the last 2 layers independently across all stages. Finally, the 2D pose module takes the $368\times 368$ image as the input and outputs $128\times 46\times 46$ 2D pose-aware feature maps for each image.

\subsection{Feature Adaption Module}
Based on the features extracted by the 2D pose module, the feature adaption module is employed to adapt the 2D pose representations into a adapted feature space for the later 3D pose prediction. As depicted in Fig.~\ref{fig:stage_detail}(b), the proposed feature adaption module consists of two convolutional layers and one fully connected layers. Each convolution layer contains 128 different kernels with the size of $5\times5$, a stride of 2, and a max pooling layer with a $2\times2$ kernel size and a stride of 2 is appended on the convolutional layers. Finally, the convolution features are fed to a fully connected layer with 1024 units to produce the adapted feature vector. In this way, the feature adapter module transforms the 2D pose-aware features into the adapted feature vector of 1024 dimensions.


\subsection{3D Pose Recurrent Module}
Given the adapted features for all frames, we propose a 3D pose sequence module to sequentially predict the 3D pose sequence. In this way, the rich temporal motion patterns between frames can be effectively incorporated into the 3D pose prediction. Note that Long Short-Term Memory (LSTM) \cite{Hochreiter1997Long} has proved better performance on exploiting temporal correlations than vanilla recurrent neural network in many tasks, \eg, speech recognition \cite{graves2014towards} and video description \cite{donahue2015long}. 
In our RPSM, the 3D pose recurrent module resorts to the LSTM layers to capture the temporal dependency in monocular sequence for refining the 3D pose prediction for each frame. 

As illustrated in Fig~\ref{fig:stage_detail}(c), the 3D pose recurrent module is constructed by one LSTM layer with 1024 hidden cells and an output layer that predicts the location of $P=17$ joint points of the human. In particular, the hidden states learned by the LSTM layers are capable of implicitly encoding the temporal dependency across different frames of the input sequence. As formulated in Eq.~(\ref{eq:opt}), the adapted features, the previous hidden states and the previous 3D pose predictions are concatenated together as the current input of 3D poses recurrent module. Incorporating the previous 3D pose prediction at each stage endows our RPSM the ability of gradually refining the pose predictions.

\section{Model Training and Testing}
In the training phase, our RPSM enforces the 3D pose sequence prediction loss for all frames at all stages, which is defined as the Euclidean distances between the prediction for all $P$ joints and ground truth:
\begin{equation}
\label{eq:loss_func}
\mathcal{L}=\sum_{k=1}^{K} \alpha_k \sum_{t=1}^{T} \left \| S_t^k - S_t^* \right \|_{2}^{2},
\end{equation}
where $K$ is the number of stages, $T$ is the length of an image sequence, $S_t^*$ is the ground-truth 3D pose for $t$-th frame, and $\alpha_k$ is the loss weight for each stage. 

The 2D Pose Module is first pretrained with MPII Human Pose dataset \cite{andriluka14cvpr}, since this dataset provides a larger variant of 2D pose data. Specifically, we temporally build up an extra convolution layer upon the public shared layers of 2D Pose Module to generate heat maps (joint confidence) as ~\cite{jt14nips}, which denote pixel-wise confidence maps of the body joints. Then we exploit the MPII Human Pose dataset \cite{andriluka14cvpr} to pretrain the tailored 2D Pose Module via the stochastic gradient decent algorithm. As for the whole framework, the ADAM~\cite{Kingma2014Adam} strategy is employed for parameter optimization .

In order to obtain sufficient samples to train the 3D pose recurrent module, we propose to decompose one long monocular image sequence into several small equal clips with $C$ frames. According to the Eq.~(\ref{eq:loss_func}), we integrally fine-tune the parameters of 3D pose recurrent module, the feature adaption module and the specialized convolutional layers of the 2D pose module in a multi-stage optimization manner. In this way, the feature adaption module can learn the adapted feature representation according to the Eq. (\ref{eq:loss_func}) for the further 3D pose estimation.

In the testing phase, every frame of the input image sequence is processed by our proposed RPSM in a stage-by-stage manner. In the end, after the final stage refinement, we output the 3D pose prediction.

\section{Experiments}
\subsection{Experimental Settings}
We perform the extensive evaluations on two publicly available datasets: Human3.6M \cite{huamn3.6m} and HumanEva-I \cite{sigal2010humaneva}. 

\begin{table*}[tp]
\setlength{\tabcolsep}{1.5pt}
	\centering
	\begin{adjustbox}{max width=1.0\textwidth}
	\begin{tabular}{@{}l|ccccccccccccccc|c@{}}
	\toprule
	 Method & Direction & Discuss & Eating & Greet & Phone & Pose & Purchase & Sitting & SitDown & Smoke & Photo & Wait & Walk & WalkDog & WalkPair & Avg. \\ \midrule
	LinKDE \cite{huamn3.6m}& 132.71 & 183.55 & 132.37 & 164.39 & 162.12 & 150.61 & 171.31 & 151.57 & 243.03 & 162.14 & 205.94 & 170.69 & 96.60 & 177.13 & 127.88 & 162.14 \\
	Li \etal \cite{li2015maximum} & - & 136.88 & 96.94 & 124.74 & - & - & - & - & - & - & 168.68 &  - & 69.97 & 132.17 & - & - \\
	Tekin \etal \cite{Tekin_2016_CVPR} & 102.39 & 158.52 & 87.95 & 126.83 & 118.37 & 114.69 & 107.61 & 136.15 & 205.65 & 118.21 & 185.02 & 146.66 & 65.86 & 128.11 & 77.21 & 125.28 \\
	Zhou \etal \cite{zhou2015sparseness} & 87.36 & 109.31 & 87.05 & 103.16 & 116.18 & 106.88 & 99.78 & 124.52 & 199.23 & 107.42 & 143.32 & 118.09 & 79.39 & 114.23 & 97.70 & 113.01 \\
	Zhou \etal \cite{zhou2016deep} & 91.83 & 102.41 & 96.95 & 98.75 & 113.35 & 90.04 & 93.84 & 132.16 & 158.97 & 106.91 & 125.22 & 94.41 & 79.02 & 126.04 & 98.96 & 107.26 \\
	Du \etal \cite{DBLP:conf/eccv/DuWLHGWKG16} & 85.07 & 112.68 & 104.90 & 122.05 & 139.08 & 105.93 & 166.16 & 117.49 & 226.94 & 120.02 & 135.91 & 117.65 & 99.26 & 137.36 & 106.54 & 126.47 \\
	Sanzari \etal \cite{DBLP:conf/eccv/SanzariNP16} & \textbf{48.82} & \textbf{56.31} & 95.98 & 84.78 & 96.47 & 66.30 & 107.41 & 116.89 & 129.63 & 97.84 & 105.58 & \textbf{65.94} & 92.58 & 130.46 & 102.21 & 93.15 \\
	Ours & 58.02 & 68.16 & \textbf{63.25} & \textbf{65.77} & \textbf{75.26} & \textbf{61.16} & \textbf{65.71} & \textbf{98.65} & \textbf{127.68} & \textbf{70.37} & \textbf{93.05} & 68.17 & \textbf{50.63} & \textbf{72.94} & \textbf{57.74} & \textbf{73.10} \\
\bottomrule
	\end{tabular}
	\end{adjustbox}
	\vspace{-10pt}
	\caption{\textbf{Quantitative comparisons on Human3.6M dataset} using 3D pose errors (in millimeter) for different actions of subjects 9 and 11. The entries with the smallest 3D pose errors for each category are bold-faced. Our RPSM achieves the significant improvement over all compared state-of-the-art approaches, \ie reduces mean error by \textbf{21.52\%}. }
	\label{table:h3m_res}
\vspace{-10pt}
\end{table*}

\textbf{Human3.6M dataset.} The Human3.6M dataset is a recently published dataset, which provides 3.6 million 3D human pose images and corresponding annotations in a controlled laboratory environment. It captures 11 professional actors performing in 15 scenarios under 4 difference viewpoints. In the following experiments, we strictly follow the same data partition protocol as in previous works
~\cite{zhou2015sparseness,li2015maximum,zhou2016deep,Tekin_2016_CVPR,DBLP:conf/eccv/DuWLHGWKG16,DBLP:conf/eccv/SanzariNP16}. The data from five subjects (S1,S5,S6,S7,S8) is for training and two subjects (S9,S11) is for testing. Note that to increase the number of training samples, the sequences from different viewpoints of the same subject are treated as distinct sequences. 
Through downsampling the frame rate from 50FPS to 2FPS, 62,437 human pose images (104 images per sequence) are obtained for training while 21,911 images for testing (91 images per sequence). To be more general, our RPSM is trained on training samples from all 15 actions instead of exploiting individual action like~\cite{zhou2015sparseness,li2015maximum}.

\begin{table*}[tp]
	\centering
	\scriptsize
	\resizebox{\textwidth}{!}{%
		\begin{tabular}{@{}l|cccc|cccc|cccc@{}}
			\toprule
			& \multicolumn{4}{c|}{Walking}                             & \multicolumn{4}{c|}{Jogging}                            & \multicolumn{4}{c}{Boxing}                                    \\ 
			Methods                  & S1            & S2            & S3            & Avg.       & S1            & S2            & S3            & Avg.       & S1            & S2            & S3            & Avg.       \\ \midrule
			Simo-Serra \etal \cite{simo2012single} & 99.6 & 108.3 & 127.4 & 111.8 & 109.2 & 93.1 & 115.8 & 108.9 & - & - & - & - \\ 
			Radwan \etal \cite{radwan2013monocular} & 75.1 & 99.8 & 93.8 & 89.6 & 79.2 & 89.8 & 99.4 & 89.5 & - & - & - & - \\
			Wang \etal \cite{wang2014robust} & 71.9 & 75.7 & 85.3 & 77.6 & 62.6 & 77.7 & 54.4 & 71.3 & - & - & - & - \\
			Du \etal \cite{DBLP:conf/eccv/DuWLHGWKG16} & 62.2 & 61.9 & 69.2 & 64.4 & 56.3 & 59.3 & 59.3 & 58.3 & - & - & - & - \\
			Simo-Serra \etal \cite{SimoSerraCVPR2013} & 65.1 & 48.6 & 73.5 & 62.4 & 74.2 & 46.6 & 32.2 & 56.7 & - & - & - & - \\
			Bo \etal \cite{bo2010twin}  & 45.4 & 28.3 & 62.3 & 45.3 & 55.1 & 43.2 & 37.4 & 45.2 & 42.5 & 64.0 & 69.3 & 58.6 \\
			Kostrikov \etal \cite{BMVC2880} & 44.0 & 30.9 & 41.7 & 38.9 & 57.2 & 35.0 & 33.3 & 40.3 & - & - & - & - \\
			Tekin \etal \cite{Tekin_2016_CVPR} & 37.5 & 25.1 & 49.2 & 37.3 & - & - & - & - & 50.5 & 61.7 & \textbf{57.5} & 56.6 \\
			Yasin \etal \cite{yasin2016dual}  & 35.8 & 32.4 & 41.6 & 36.6 & 46.6 & 41.4 & 35.4 & 38.9 & - & - & - & - \\
			Ours & \textbf{26.5} & \textbf{20.7} & \textbf{38.0} & \textbf{28.4} & \textbf{41.0} & \textbf{29.7} & \textbf{29.1} & \textbf{33.2} & \textbf{39.4} & \textbf{57.8} & 61.2 & \textbf{52.8} \\ \bottomrule
		\end{tabular}%
	}
	\vspace{-10pt}
	\caption{\textbf{Quantitative comparisons on HumanEva-I dataset} using 3D pose errors (in millimeter) for the ``Walking", ``Jogging" and ``Boxing" sequences. '-' indicates the corresponding method has not reported the accuracy on that action. The entries with the smallest 3D pose errors for each category are bold-faced. Our RPSM outperforms all the compared state-of-the-art methods by a clear margin. }
	\label{table:hevai_res}
	\vspace{-12pt}
\end{table*}

\textbf{HumanEva-I dataset.} The HumanEva-I dataset contains video sequences of four subjects performing six common actions(\eg, walking, jogging, boxing \etc), and it also provides the 3D pose annotation for each frame in the video sequences. We train our RPSM on training sequences of the subject 1, 2 and 3 and test on the `validation' sequence in the same protocol as \cite{yasin2016dual,Tekin_2016_CVPR,simo2012single,SimoSerraCVPR2013,BMVC2880,bo2010twin,radwan2013monocular,wang2014robust}. 
Similar as the Human3.6M dataset, the data from different camera viewpoints is also regarded as different training samples. Note that we have not downsampled the video sequences to obtain more samples for training.

\textbf{Evaluation metric.} Following ~\cite{zhou2015sparseness,DBLP:conf/eccv/DuWLHGWKG16,Tekin_2016_CVPR}, we employ the popular \emph{3D pose error} metric~\cite{simo2012single} , which calculates the Euclidean errors on all joints and all frames up to translation. In the following section, we will report the 3D pose error metric for all the experimental comparisons and analysises.

\begin{figure*}
\center
\includegraphics[width= 0.9 \textwidth]{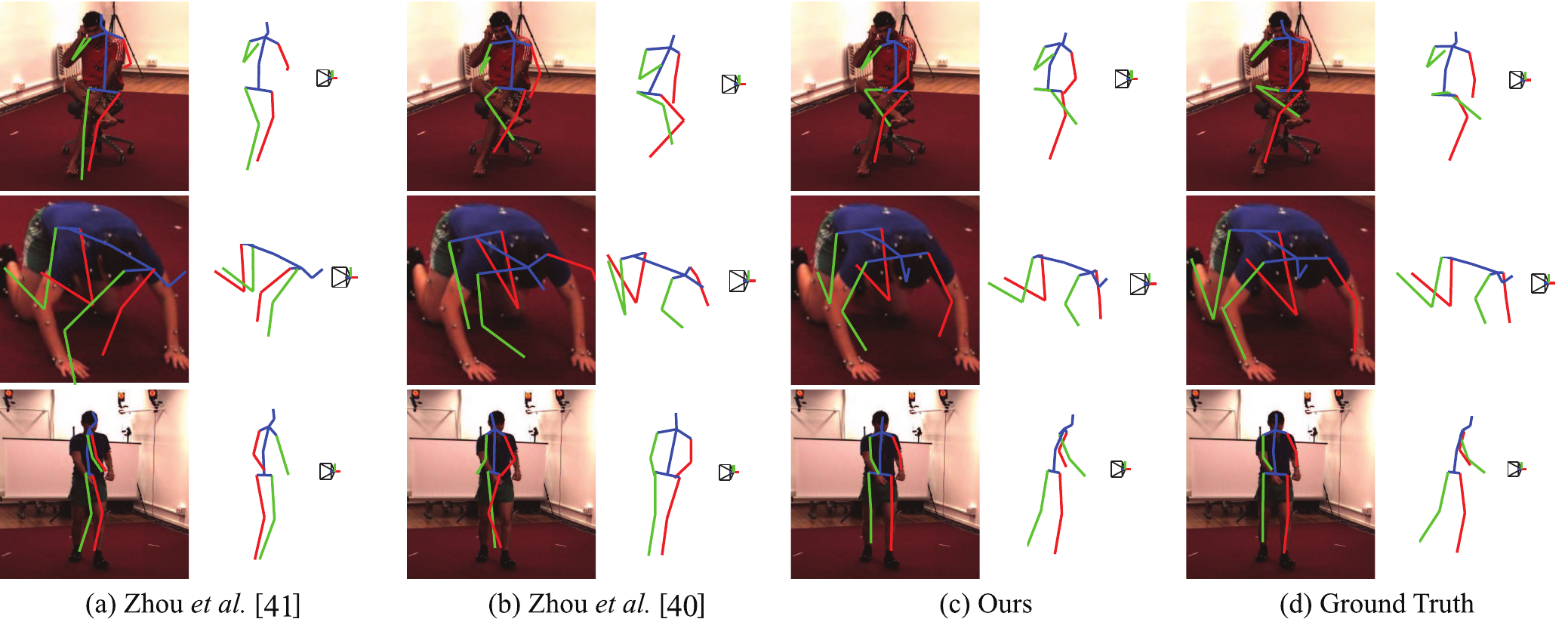}
\vspace{-5pt}
	\caption{Empirical study on the qualitative comparisons on Human3.6M dataset. The 3D pose are visualized from the side view and the camera are also depicted. Zhou \etal \cite{zhou2015sparseness}, Zhou \etal \cite{zhou2016deep}, our RPSM and the ground truth are illustrated from left to right, respectively. Our RPSM achieves much more accurate estimations than the methods of Zhou \etal \cite{zhou2015sparseness} and Zhou \etal \cite{zhou2016deep}. Best view in color.}
	\vspace{-8pt}
	\label{fig:h3m_vis}
\end{figure*}

\begin{table*}[htpb]
	\centering
	\setlength{\tabcolsep}{1.5pt}
	\begin{adjustbox}{max width=1.0\textwidth}
	\begin{tabular}{@{}c|ccccccccccccccc|c@{}}
	\toprule 
	Method & Direction & Discuss & Eating & Greet & Phone & Pose & Purchase & Sitting & SitDown & Smoke & Photo & Wait & Walk & WalkDog & WalkPair & Avg. \\ \midrule
	RPSM-1-stage & 62.89 & 74.74 & 67.86 & 73.33 & 79.76 & 67.48 & 76.19 & 100.21 & 148.03 & 75.95 & 100.26 & 75.82 & 58.03 & 78.74 & 62.93 & 80.15 \\
	RPSM-2-stage & 58.96 & 68.50 & 65.64 & 68.18 & 78.41 & 62.82 & 67.04 & 100.63 & 136.72 & 73.35 & 96.87 & \textbf{67.96} & 51.64 & 77.27 & 59.31 & 75.55 \\
	RPSM-3-stage & \textbf{58.02} & \textbf{68.16} & \textbf{63.25} & \textbf{65.77} & \textbf{75.26} & \textbf{61.16} & \textbf{65.71} & \textbf{98.65} & \textbf{127.68} & \textbf{70.37} & \textbf{93.05} & 68.17 & \textbf{50.63} & \textbf{72.94} & \textbf{57.74} & \textbf{73.10} \\ 
	RPSM\_1stage\_seq\_1 & 70.46 & 83.36 & 76.46 & 80.96 & 88.14 & 76.00 & 92.39 & 116.62 & 163.14 & 85.87 & 111.46 & 83.60 & 65.38 & 95.10 & 73.54 & 90.83 \\
	RPSM\_3stage\_seq\_1 & 61.94 & 75.84 & 65.25 & 71.28 & 79.39 & 67.73 & 77.88 & 105.47 & 153.58 & 76.01 & 101.84 & 74.12 & 56.07 & 85.63 & 64.78 & 81.12 \\
	\hline
		RPSM\_1stage\_seq\_5 & \textbf{62.89} & \textbf{74.74} & \textbf{67.86} & \textbf{73.33} & \textbf{79.76} & \textbf{67.48} & 76.19 & \textbf{100.21} & 148.03 & \textbf{75.95} & \textbf{100.26} & \textbf{75.82} & \textbf{58.03} & \textbf{78.74} & \textbf{62.93} & \textbf{80.15} \\
	RPSM\_1stage\_seq\_10 & 66.73 & 76.82 & 73.57 & 76.56 & 84.80 & 70.57 & \textbf{75.44} & 110.70 & \textbf{143.10} & 80.35 & 103.61 & 75.66 & 58.52 & 80.55 & 66.19 & 82.88 \\ 
	\hline
		RPSM-3-stage\_no\_MPII & 91.58 &109.35&93.28&98.52&102.16&93.87&118.15&134.94	&190.6&109.39&121.49&101.82&88.69&110.14&105.56&111.3 \\
		RPSM-3-stage\_sharing & \textbf{58.36} & \textbf{66.52} & \textbf{63.37} & \textbf{64.5} & \textbf{72.22} & \textbf{59.39} & \textbf{63.9} & \textbf{90.73} & \textbf{129.99} &	\textbf{68.26} & \textbf{93.86} & \textbf{65.22} & \textbf{48.47} & \textbf{70.53} & \textbf{56.26} & \textbf{71.44} \\

	\bottomrule
	\end{tabular}
	\end{adjustbox}
	\vspace{-10pt}
	\caption{Top five rows: empirical study on different number of refinement stages. Middle two rows: empirical comparisons by different sequence lengths (i.e., 1, 5, 10) for each clip. Note that the results are evaluated by a single-stage RPSM. Bottom two rows: performance of RPSM variants. The entries with the smallest 3D pose errors on Human3.6m dataset for each category are bold-faced.}
	\vspace{-12pt}
	\label{table:h3m_across_stages}
	\label{tab:rpsm-rho3-ext}
	\label{table:h3m_across_time}
\end{table*}

%
%
%


\begin{figure*}[h]
\centering
\includegraphics[width=0.95 \textwidth]{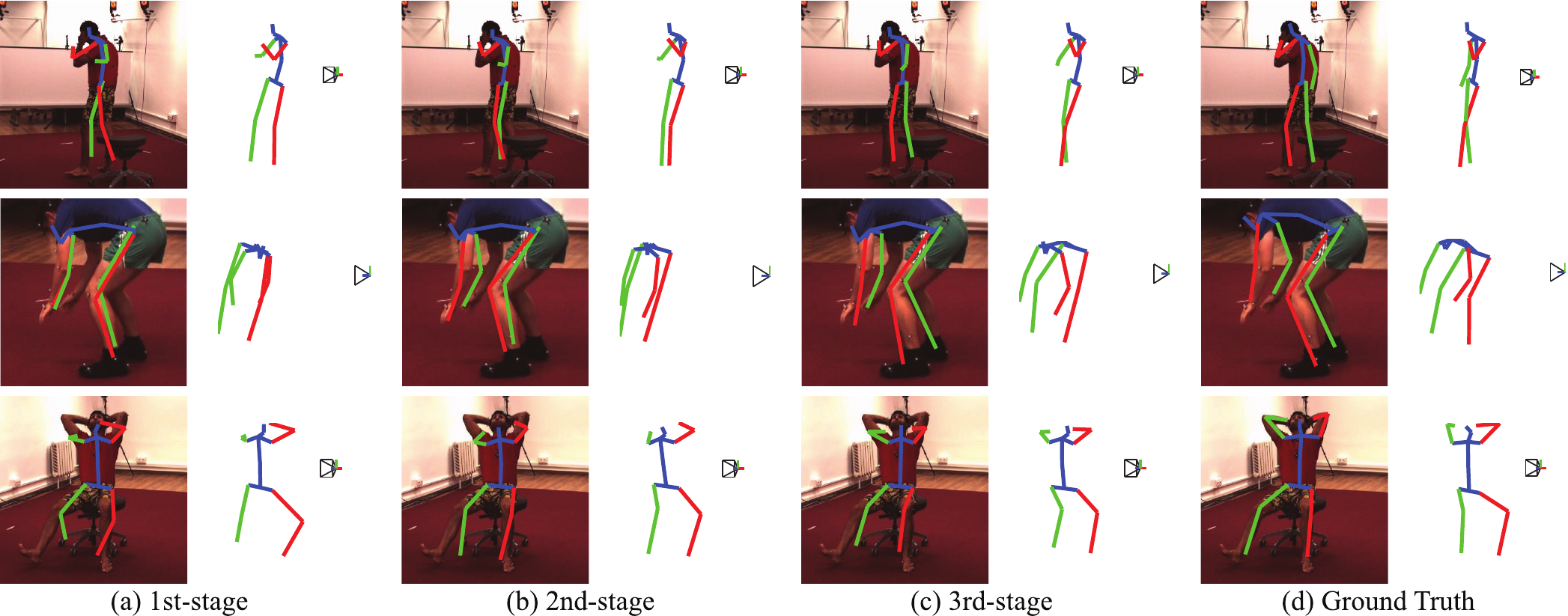}
\vspace{-8pt}
\caption{Qualitative comparisons of different stage refinement on Human3.6M dataset. The estimated 3D skeletons are reprojected into the images and shown by themselves from the side view (next to the images). The figures from left to right correspond to the estimated 3D poses generated by the 1st-stage, 2nd-stage, 3rd-stage of our RPSM and ground truth, respectively. We can observe that the predicted human 3D joints are progressively corrected along with the multi-stage sequential learning. Best viewed in color.}
\vspace{-15pt}
\label{fig:3d_pose_example2}
\end{figure*}

\textbf{Implementation Details:}
Our RPSM is implemented by using Torch7 \cite{collobert2011torch7} deep learning toolbox. We follow \cite{graves2014towards} to build the LSTM memory cells, except that the peephole connections between cell and gates are omitted. The loss weights $\alpha_k$ for each stage are all set to 1. In total, three-stage refinements are performed for all our experiments since only unnoticeable performance difference is observed using more stages. 
Following~\cite{zhou2015sparseness,li2015maximum}, the input image is cropped around the human. To keep the human ratio, we crop a square image of the subject from the image according to the bounding box provided by the dataset. 
Then we resize the image region inside the bounding box into 368$\times$368 resolution before feeding it into the network. Moreover, we augment the training data only by random scaling with factors in [0.9,1.1]. Note that to transform the absolute locations of joint points into the [0,1] range, $\text{max-min}$ normalization strategy is applied. In the testing phase, the predicted 3D pose is transform to the origin scale according to the maximum and minimal value of the pose from training frames. During the training, the Xavier initialization method \cite{glorot2010understanding} was used to initialize the weights of our RPSM. The decay is set as $1e^{-4}$ and base learning rate of $1e^{-3}$ is employed for training. It took about 2 days to train a 3-stage RPSM with 50 epochs on single NVIDIA GeForce GTX TITAN X with 12GB memory. In the testing phase, it takes about 50 ms to process an image.

\subsection{Comparisons with state-of-the-art methods}

\textbf{Comparison on Human3.6M:}
\label{section:human3.6m_result}
We compare our RPSM with the state-of-the art methods on Human3.6M \cite{huamn3.6m} and HumanEva-I \cite{sigal2010humaneva} dataset. These state-of-the-art methods are LinKDE~\cite{huamn3.6m}, Tekin \etal \cite{Tekin_2016_CVPR}, Li \etal \cite{li2015maximum}, Zhou \etal \cite{zhou2015sparseness} (CNN based), Zhou \etal \cite{zhou2016deep}, Du \etal \cite{DBLP:conf/eccv/DuWLHGWKG16} and Sanzari \etal \cite{DBLP:conf/eccv/SanzariNP16}.

The results are summarized in Table \ref{table:h3m_res}. As one can see from Table \ref{table:h3m_res}, our proposed RPSM model significantly outperforms all compared methods with mean error reduced by \textbf{31.85\%} compared with \cite{zhou2016deep} and \textbf{21.52\%} compared with \cite{DBLP:conf/eccv/SanzariNP16}. Note that some compared methods, \eg, \cite{li2015maximum,Tekin_2016_CVPR,DBLP:conf/eccv/DuWLHGWKG16,zhou2015sparseness,zhou2016deep}, also employ deep learning techniques. Especially, Zhou \etal \cite{zhou2016deep}'s method has used the recently published Residual Network \cite{he2015deep}. This superior performance achieved by RPSM demonstrates that utilizing multi-stage RPSM is simple yet powerful in capturing complex contextual features within images and learning temporal dependency within image sequences, which are critical for estimating 3D pose sequence.

\textbf{Comparison on HumanEva-I:} 
On this dataset, we compare our RPSM against methods which rely on several kinds of separate processing steps. These methods include discriminative regressions\cite{bo2010twin,BMVC2880}, 2D pose detectors based \cite{simo2012single,SimoSerraCVPR2013,wang2014robust,yasin2016dual}, CNN-based regressions \cite{Tekin_2016_CVPR}. 
For fair comparison, our RPSM also predicts the 3D pose consisting of 14 joints, i.e., left/right shoulder, elbow, wrist, left/right hip knee, ankle, head top and neck, as \cite{yasin2016dual}. 

Table \ref{table:hevai_res} illustrates the performance comparisons between our RPSM with compared methods. It is obvious that our RPSM model obtains substantially lower 3D pose errors than the compared methods, and achieves new state-of-the-art performance on all \emph{Walking, Jogging and Boxing} sequences. In addition, in terms of the time efficiency, compared with \cite{bo2010twin} which takes around three minutes per image and \cite{yasin2016dual} which takes more than 25 seconds per image, our RPSM model only costs 50ms per image. This demonstrate the effectiveness and efficiency of our proposed RPSM model.

\subsection{Component Analysis}

\textbf{Effectiveness of multi-stage refinement:} 
To validate the superiority of the proposed multi-stage refinement of our RPSM, we conduct the following experiment: employing one, two, three stages for human pose estimation and denote them as ``RPSM-1-stage'', ``RPSM-2-stage'' and ``RPSM-3-stage''. The evaluations are performed on the Human3.6M dataset from the qualitative and quantitative aspects. The top five rows of Table~\ref{table:h3m_across_stages} illustrates the comparisons of estimating 3D pose errors for using different number of stages. As one can see from Table~\ref{table:h3m_across_stages}, the performance increases monotonically within 3 stages. Moreover, the single/multi-stage performance without temporal dependency 
is also compared in Table.~\ref{tab:rpsm-rho3-ext} (denoted as ``RPSM-1stage\_seq\_1'' and ``RPSM-3stage\_seq\_1'', respectively). As illustrated in Table.~\ref{tab:rpsm-rho3-ext}, RPSM-3stage\_seq\_1 has achieved much lower 3D pose errors than RPSM-1stage\_seq\_1 (81.12 vs 90.83). This validates that the effectiveness of multi-stage refinement even when temporal information is ignored. Thanks to the exploited richer contextual information, our RPSM can learn more robust 2D pose-aware features and the representation of 3D pose sequences. Exemplar visual results on three different stages are shown in Fig.~\ref{fig:3d_pose_example2}. It can be seen that the joint predictions are progressively corrected by performing multi-stage refinement.

\textbf{Pre-training and Weight Sharing:}
To evaluate the performance without pre-training, we have only employed Human3.6m 2D pose data and annotations to train the 2D pose module. We denote this version of our RPSM as ``RPSM-3-stage\_no\_MPII''. The result is reported in the bottom two rows of Table.~\ref{tab:rpsm-rho3-ext}.  As one can see from Table.~\ref{tab:rpsm-rho3-ext}, RPSM-3-stage\_no\_MPII performs quite worse than RPSM-3-stage. This may be due to that Human3.6m 2D pose data, compared with MPII dataset, is less challenging for CNN to learn a rich 2D pose presentation. 
Note that according to the bottom row of Table.~\ref{tab:rpsm-rho3-ext}, the performance of sharing all layers in the 2D pose module (denoted as ``RPSM-3stage-sharing'') is slightly better than the partially sharing one (denoted as ``RPSM-3-stage''). However, the training time will significantly increased. Thus, we decide to choose partially sharing manner.

\textbf{Importance of temporal dependency:} To study the effectiveness of incorporating temporal dependency, we also evaluate the variants of our single-stage RPSM using different clip lengths, i.e., 1, 5 and 10, named as ``RPSM-1stage-seq\_$C$", where $C$ denotes the frame number of clips for training. Note that, when $C$ is equal to 1, no temporal information is considered and thus the recurrent LSTM layer in 3D pose errors is replaced with a fully connected layer with the same units as the LSTM. 
Results of using different clip length are reported in Table \ref{table:h3m_across_time}. From the comparison results, the importance of temporal dependency is well demonstrated. Considering temporal dependency methods (i.e., RPSM\_1stage\_seq\_5 and RPSM\_1stage\_seq\_10) all outperform the RPSM\_1stage\_seq\_1 in a clear margin (about 10\% reduction of the mean joint errors on the Human3.6M dataset). The minor performance difference between RPSM\_1stage\_seq\_5 and RPSM\_1stage\_seq\_10 may be due to effect of temporal inconsistency, which has higher probability to occur in long clips. 
Moreover, it also should be noted that ``RPSM\_1stage\_seq\_1" shows superiority over all state-of-the-art approaches owing to the contribution of the proposed 2D pose module and feature adaption module.

\section{Conclusion}
We have proposed a novel Recurrent 3D Pose Sequence Machines (RPSM) for estimating 3D human pose from a sequence of monocular images. Through the proposed unified architecture with 2D pose, feature adaption and 3D pose recurrent modules, our RPSM can learn to recurrently integrate rich spatio-temporal long-range dependencies in an implicit and comprehensive way. We also proposed to employ multiple sequential stages to refine the estimation results via the 3D pose geometry information. The extensive evaluations on two public 3D human pose dataset validate the effectiveness and superior performance of the our RPSM. In future work, we will extend the proposed framework for other sequence-based human centric analysis such as human action and activity recognition.

{\small
\bibliographystyle{ieee}
\bibliography{human_pose}
}

\end{document}